\documentclass[letterpaper, 10 pt, conference]{ieeeconf}  

\IEEEoverridecommandlockouts                              

\usepackage{graphics} 
\usepackage{epsfig} 
\usepackage{times} 
\usepackage{amsmath} 
\usepackage{amssymb}  

\makeatletter
\let\NAT@parse\undefined
\makeatother

\usepackage[numbers]{natbib}

\usepackage{url}
\usepackage{pifont}
\newcommand{\cmark}{\ding{51}}%
\newcommand{\xmark}{\ding{55}}%

\usepackage{soul}
\sethlcolor{yellow}
\soulregister\ref7
\soulregister\circ7

\usepackage[utf8]{inputenc} 
\usepackage[T1]{fontenc}    
\usepackage{url}            
\usepackage{booktabs}       
\usepackage{amsfonts}       
\usepackage{nicefrac}       
\usepackage{microtype}      
\usepackage{multirow}
\usepackage{graphicx} 
\usepackage[pagebackref=true,breaklinks=true,colorlinks,bookmarks=false]{hyperref}
\hypersetup{citecolor=[RGB]{119,185,0}}

\usepackage{balance}
\newcommand\crule[3][black]{\textcolor{#1}{\rule{#2}{#3}}}
\usepackage{xcolor}
\definecolor{nvcolor}{RGB}{119,185,0}
\definecolor{roadcolor}{RGB}{234,51,246}
\definecolor{sidewalkcolor}{RGB}{68,8,72}
\definecolor{parkingcolor}{RGB}{241,156,249}
\definecolor{othergroundcolor}{RGB}{160,32,76}
\definecolor{buildingcolor}{RGB}{246,202,69}
\definecolor{carcolor}{RGB}{111,149,238}
\definecolor{truckcolor}{RGB}{74,32,172}
\definecolor{bicyclecolor}{RGB}{136,227,242}
\definecolor{motorcyclecolor}{RGB}{37,59,146}
\definecolor{othervehiclecolor}{RGB}{96,81,242}
\definecolor{vegetationcolor}{RGB}{79, 173, 50}
\definecolor{trunkcolor}{RGB}{126, 65, 22}
\definecolor{terraincolor}{RGB}{171, 238, 105}
\definecolor{personcolor}{RGB}{234, 60, 49}
\definecolor{bicyclistcolor}{RGB}{234, 66, 195}
\definecolor{motorcyclistcolor}{RGB}{138, 42, 90}
\definecolor{fencecolor}{RGB}{238, 128, 69}
\definecolor{polecolor}{RGB}{252, 241, 161}
\definecolor{trafficsigncolor}{RGB}{233, 51, 35}
\definecolor{other-struct.color}{RGB}{255, 150, 0}
\definecolor{other-objectcolor}{RGB}{50, 255, 255}
\definecolor{lane-markingcolor}{RGB}{150, 255, 170}
\definecolor{color1}{RGB}{176, 36, 24}
\definecolor{color2}{RGB}{0, 176, 80}
\definecolor{color3}{RGB}{0, 0, 200}
\definecolor{colorofteaser}{RGB}{176, 36, 24}

\usepackage{colortbl}
\definecolor{mygray}{gray}{0.9}
\newcommand{\tbr}[1]{\textbf{\textcolor{color1}{#1}}}
\newcommand{\tbg}[1]{\textbf{\textcolor{color2}{#1}}}
\newcommand{\tbb}[1]{\textbf{\textcolor{color3}{#1}}}

\title{\LARGE \bf
SSCBench: A Large-Scale 3D Semantic Scene Completion \\Benchmark for Autonomous Driving
}

\author{Yiming Li$^{1, *}$, Sihang Li$^{1, *}$, Xinhao Liu$^{1, *}$, Moonjun Gong$^{1, *}$, {Kenan Li}$^{1}$, {Nuo Chen}$^{1}$, \\ {Zijun Wang}$^1$, {Zhiheng Li}$^1$, {Tao Jiang}$^2$, {Fisher Yu}$^{3}$, {Yue Wang}$^{4}$, {Hang Zhao}$^{2}$,  {Zhiding Yu}$^{4}$, {Chen Feng}$^{1,\dagger}$ 
\thanks{$^{1}$New York University, $^2$Tsinghua University, $^{3}$ETH Zurich, $^4$NVIDIA}%
\thanks{$^*$Equal contribution. Yiming Li is supported by NVIDIA Fellowship.}%
\thanks{$^\dagger$The corresponding author is Chen Feng {\tt\small cfeng@nyu.edu}. The work was supported by NSF 2238968 and 2322242 grants; and in part through the NYU IT High Performance Computing resources, services, and staff expertise.}
}

\begin{document}

\maketitle
\thispagestyle{empty}
\pagestyle{empty}

\begin{abstract}

Monocular scene understanding is a foundational component of autonomous systems. Within the spectrum of monocular perception topics, one crucial and useful task for holistic 3D scene understanding is semantic scene completion (SSC), which jointly completes semantic information and geometric details from RGB input. However, progress in SSC, particularly in large-scale street views, is hindered by the scarcity of high-quality datasets. To address this issue, we introduce SSCBench, a comprehensive benchmark that integrates scenes from widely used automotive datasets (e.g., KITTI-360, nuScenes, and Waymo). SSCBench follows an established setup and format in the community, facilitating the easy exploration of SSC methods in various street views. We benchmark models using monocular, trinocular, and point cloud input to assess the performance gap resulting from sensor coverage and modality. Moreover, we have unified semantic labels across diverse datasets to simplify cross-domain generalization testing. We commit to including more datasets and SSC models to drive further advancements in this field. Our data and code are available at \url{https://github.com/ai4ce/SSCBench}.

\end{abstract}

\section{INTRODUCTION}

Understanding 3D scenes from a single RGB image is crucial and meaningful in vision and robotics, with monocular perception tasks like object detection~\citep{wang2022monocular}, tracking~\citep{hu2022monocular}, and depth estimation~\citep{yuan2022newcrfs} garnering significant attention. The emerging field of 3D semantic scene completion (SSC)~\citep{roldao20223d} seeks to jointly infer complete 3D semantics and geometry from a sparse and partial observation (\textit{e.g.}, an RGB image). The resulting volumetric representation seamlessly integrates occupancy and semantic information, facilitating robotic scene understanding and planning capabilities in street views.

One critical challenge in SSC is to generate accurate ground truth labels, especially in street views. Given the current limitations of 3D sensing technology, achieving a perfectly comprehensive 3D representation is impossible. The pioneering SemanticKITTI benchmark~\citep{behley2019semantickitti} proposes to leverage the temporal information through the aggregation of different LiDAR sweeps, which can effectively reveal previously occluded 3D surfaces. Meanwhile, it excludes 3D voxels not observed from all viewpoints during driving. Consequently, SemanticKITTI provides relatively comprehensive and accurate ground truth labels for SSC tasks.

While SemanticKITTI is a valuable resource for learning sparse-to-dense mapping, its limited scale and diversity impede the development of more powerful and generalizable SSC models. Another significant limitation of SemanticKITTI is the omission of dynamic objects during ground truth generation, resulting in inaccurate labels. Hence, there is an urgent need for a large-scale SSC dataset with reliable ground truth to advance learning-based scene understanding in street views.

\begin{figure}[t]
  \centering
   \includegraphics[width=0.982\linewidth]{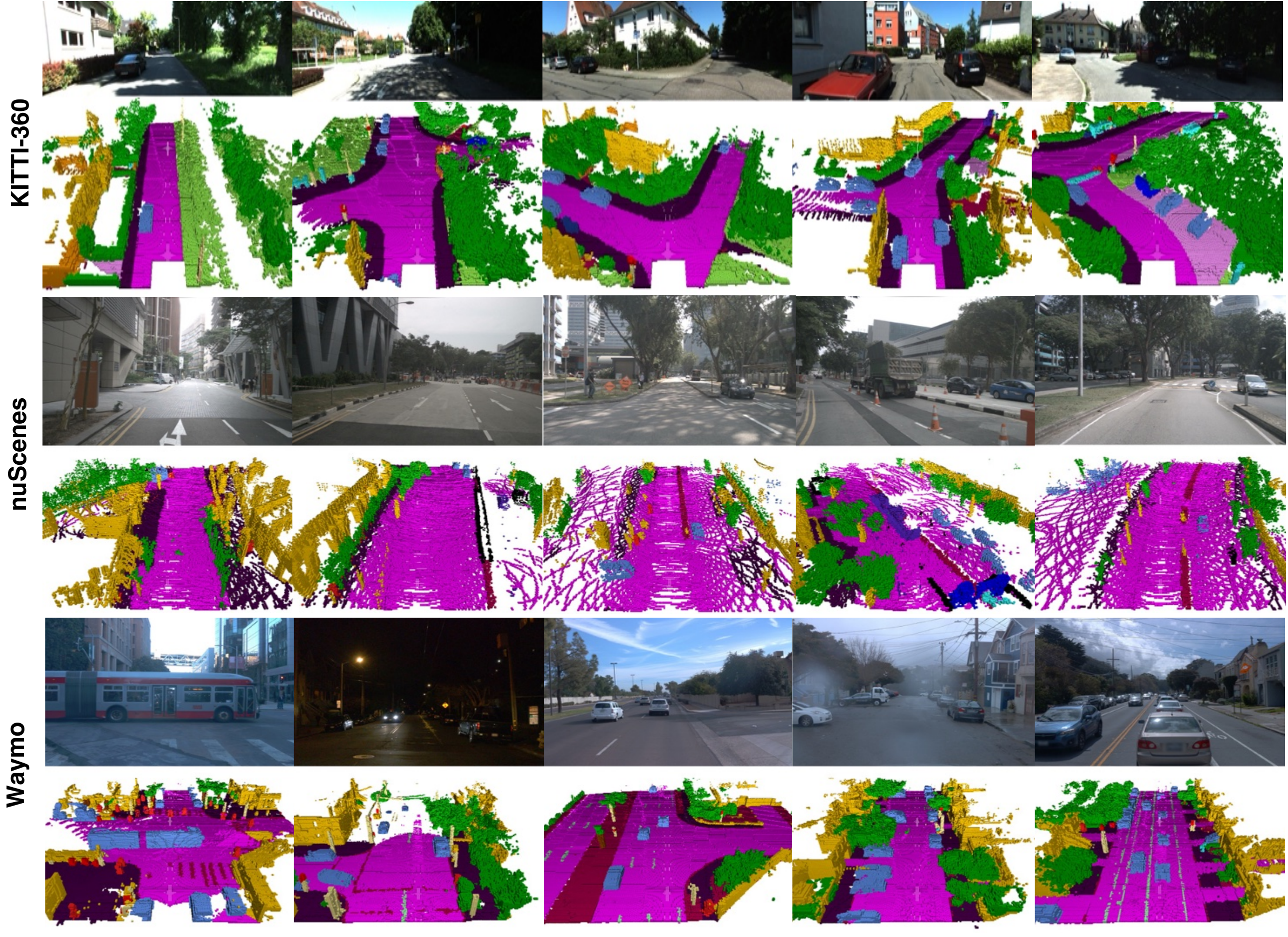}
   \vspace{-3mm}
   \caption{\textbf{Visualizations of SSCBench} derived from KITTI-360~\citep{liao2022kitti}, nuScenes~\citep{caesar2020nuscenes}, and Waymo~\citep{sun2020scalability}. We showcase accurate SSC ground truth in a variety of street views.}
   \label{fig:teaser}
   \vspace{-4mm}
\end{figure}

To this end, we introduce SSCBench, a large-scale benchmark comprising diverse street views sourced from well-established automotive datasets, including KITTI-360~\citep{liao2022kitti}, nuScenes~\citep{caesar2020nuscenes}, and Waymo~\citep{sun2020scalability}, as illustrated in Fig.~\ref{fig:teaser}. To enhance label accuracy, we utilize the 3D bounding box labels provided in these datasets to synchronize measurements of dynamic objects. Our key features include \textcolor{black}{\textbf{(a)}} \textbf{\textit{\textcolor{black}{accessibility}}}: we provide datasets in a format compatible with SemanticKITTI, facilitating seamless usage within the community; \textcolor{black}{\textbf{(b)}} \textbf{\textit{\textcolor{black}{large scale}}}: we offer an extensive dataset with $\sim$8 times more frames than SemanticKITTI, encompassing diverse geographic locations across six cities; \textcolor{black}{\textbf{(c)}} \textbf{\textit{\textcolor{black}{comprehensiveness}}}: we mainly focus on SSC methods with \textit{monocular input}. Additionally, we utilize \textit{trinocular input} to compare the single-view and panoramic-view methods and use \textit{point cloud input} to show the gap between camera-based and LiDAR-based methods. Furthermore, we have unified semantic labels across different datasets in SSCBench, facilitating cross-domain generalization experiments. We plan to continually incorporate novel automotive datasets and SSC algorithms to drive further advancements in the field.

\section{RELATED WORKS}

\textbf{Monocular Perception and 3D Semantic Scene Completion.} The simplicity, efficiency, affordability, and accessibility of monocular cameras have made monocular perception a focal point of attention in the vision and robotics community. This has resulted in extensive research into various tasks, including depth estimation~\citep{dong2022towards}, 3D object detection and tracking~\citep{enzweiler2008monocular}, as well as localization and mapping~\citep{younes2017keyframe}. \citet{song2017semantic} introduce the concept of monocular 3D semantic scene completion (SSC), which seeks to reconstruct and complete the semantics as well as geometry within a 3D volume from a single depth image. However, they only consider the bounded indoor scenarios due to the lack of outdoor datasets. \citet{behley2019semantickitti} build the first outdoor dataset based on KITTI~\citep{geiger2012we} for 3D semantic scene completion in street views. Existing approaches usually depend on 3D inputs, such as LiDAR point clouds~\citep{roldao2020lmscnet,cheng2021s3cnet,rist2021semantic}, while recent monocular vision-based solutions also emerge~\citep{cao2022monoscene,li2023voxformer}. However, the development of outdoor SSC is hindered by the lack of datasets, with SemanticKITTI~\citep{behley2019semantickitti} being the only dataset supporting SSC in street views.  Building diverse datasets is imperative to unlock the full potential of SSC for autonomous systems.

\textbf{Point Cloud Segmentation in Street Views.}
3D LiDAR segmentation aims to assign point-wise semantic labels for point clouds, including a range of specific tasks, like LiDAR semantic~\citep{hu2020randla, qiu2021semantic, milioto2019rangenet++, xu2020squeezesegv3}, panoptic~\citep{zhou2021panoptic, hong2021lidar, behley2021benchmark}, and 4D panoptic segmentation~\citep{kreuzberg20224d, aygun20214d}. In this field, point-based methods, stemming from PointNet++~\citep{qi2017pointnet}, perform well on small synthetic point cloud~\citep{yi2016scalable} rather than sparse LiDAR point cloud, with sampling and gathering disordered neighbors. Voxel-based approaches~\citep{zhu2021cylindrical, tang2020searching, li2022self, cheng20212} process point clouds by initially partitioning 3D space into voxels through Cartesian coordinates. Note that 3D LiDAR segmentation aims to understand the scenes based on raw LiDAR scans, while 3D semantic scene completion includes the completion of occluded areas, with the input of camera or LiDAR. 

\begin{table}[t]
    \centering
    \caption{\textbf{Overview of widely-used autonomous driving datasets with multimodal sensors.} C denotes camera and L denotes LiDAR. Most datasets provide bounding annotations for 3D detection, yet only a few of them provide semantic labels for the LiDAR point cloud due to the high cost. Note that ApolloScape~\citep{huang2019apolloscape} only provides 3D semantic labels for the static environments.}
    \label{tab:datasets}
    \resizebox{0.95\columnwidth}{!}{%
    \begin{tabular}{c|c|c|c|c|c}
        \toprule
        Datasets & Year & Sensors & Annotations & \# Fr. with Pts Ann.  & Sequential  \\
        \midrule
        \rowcolor{mygray} KITTI~\cite{geiger2012we} & CVPR 2012 & C\&L & 3D Bbox & N.A.  & \xmark  \\
        \rowcolor{mygray} SemanticKITTI~\cite{behley2019semantickitti} & ICCV 2019 & C\&L  & 3D Pts. & 20K  & \cmark  \\\midrule
        \rowcolor{mygray} {nuScenes}~\cite{caesar2020nuscenes} & CVPR 2019 & C\&L &  3D Bbox & N.A.  & \cmark \\
        \rowcolor{mygray} \rowcolor{mygray} {Panoptic nuScenes}~\cite{fong2022panoptic} & RA-L 2022 & C\&L &  3D Pts. & 40K & \cmark \\ \midrule
        \rowcolor{mygray} {Waymo}~\cite{sun2020scalability} & CVPR 2020 & C\&L & 3D Bbox\&Pts. & 230K & \cmark \\\midrule
        \rowcolor{mygray} {KITTI-360}~\cite{liao2022kitti} & T-PAMI 2022 & C\&L & 3D Bbox\&Pts. & 100K  & \cmark \\ \midrule
        ApolloScape~\cite{huang2019apolloscape} & T-PAMI 2019 & C\&L & 3D Bbox\&Pts.  & N.A.  & \cmark \\ \midrule
        Argoverse~\cite{chang2019argoverse} & CVPR 2019 & C\&L &  3D Bbox & N.A. & \cmark \\ \midrule
        ONCE~\cite{mao2021one} & NeurIPS 2021 & C\&L &  3D Bbox & N.A. & \cmark \\ \midrule
        Lyft Level 5~\cite{WovenPlanetHoldingsInc2019} & 2019 & C\&L &  3D Bbox & N.A. & \cmark  \\ \midrule
        A*3D~\cite{pham20203d} & ICRA 2020 & C\&L &  3D Bbox & N.A. & \cmark \\ \midrule
        A2D2~\cite{geyer2020a2d2} &  2020 & C\&L & 3D Bbox& N.A. & \xmark \\ 
        \bottomrule
    \end{tabular}
    }
\end{table}

\textbf{Autonomous Driving Dataset and Benchmark.} 
Autonomous driving research thrives on high-quality datasets, which serve as the lifeblood for training and evaluating perception~\citep{caesar2020nuscenes}, prediction~\citep{ettinger2021large}, and planning algorithms~\citep{hu2023planning}. In 2012, the pioneering KITTI dataset sparked a revolution in autonomous driving research, unlocking a multitude of tasks including object detection, tracking, mapping, and optical/depth estimation~\citep{geiger2012we,Geiger2013IJRR,Fritsch2013ITSC,Menze2015CVPR,chen2023deepmapping2}. Since then, the research community has embraced the challenge, giving rise to a wealth of datasets. These datasets push the boundaries of autonomous driving research by addressing challenges posed by multimodal fusion~\citep{caesar2020nuscenes}, multi-tasking learning~\citep{huang2018apollo, liao2022kitti}, adverse weather~\citep{pitropov2021canadian}, collaborative driving~\citep{agarwal2020ford,li2022v2x,xu2023v2v4real}, repeated driving~\citep{diaz2022ithaca365}, and dense traffic scenarios~\citep{pham20203d,xiao2021pandaset}, \textit{etc}.
There are several impactful and widely-used driving datasets such as KITTI-360~\citep{liao2022kitti}, nuScenes~\citep{caesar2020nuscenes}, and Waymo~\citep{sun2020scalability}. They provide LiDAR and camera recordings as well as point cloud semantics and bounding annotations, as summarized in Tab.~\ref{tab:datasets}. Therefore, we can create accurate ground truth labels for SSC by aggregating multiple semantic point clouds and leveraging the 3D boxes to handle dynamic objects.

\textbf{Occ3D and OpenOccupancy.} We compare SSCBench with the concurrent relevant work Occ3D~\citep{tian2023occ3d}. The differences lie in: \textbf{\textit{(a) setup}}: Occ3D uses surrounding-view images as input, and only considers the reconstruction of 3D voxels visible to the camera. SSCBench considers a more challenging yet meaningful setup (also a well-established one): how to reconstruct and complete 3D semantics in both visible and occluded areas only with monocular visual input. \textit{This task requires reasoning about temporal information and 3D geometric relationships to get rid of the limited field of view}; \textbf{\textit{(b) scale}}: SSCBench provides more datasets than Occ3D and plans to add more due to the abundance of monocular driving recordings; \textbf{\textit{(c) accessibility}}: we inherit the widely-used setup from the pioneer KITTI, thus making SSCBench more accessible to the community; \textbf{\textit{(d) comprehensiveness}}: we benchmark SSC methods with monocular, trinocular, and point cloud input and provide unified labels for cross-domain generalization tests. Another relevant benchmark, OpenOccupancy~\citep{wang2023openoccupancy}, exhibits similar differences, notably its exclusive use of the nuScenes dataset~\citep{caesar2020nuscenes}, which results in a limitation of diversity.

\section{Dataset Curation}

\subsection{Revisit of SemanticKITTI}

SemanticKITTI~\citep{behley2019semantickitti} extends the odometry dataset of the KITTI vision benchmark~\citep{geiger2012we} by providing point-wise semantic annotations for 22 driving sequences in Karlsruhe, Germany. SemanticKITTI not only supports 3D semantic segmentation but also serves as the first outdoor SSC benchmark. Similar to indoor SSC~\citep{song2017semantic}, SemanticKITTI uses voxelized 3D representation widely employed in robotics such as occupancy grid mapping~\citep{thrun2002probabilistic}. SemanticKITTI generates ground truth labels via voxelization of a dense semantic point cloud given by rigid registration of multiple LiDAR scans.

\begin{figure}[t]
  \centering
   \includegraphics[width=0.95\linewidth]{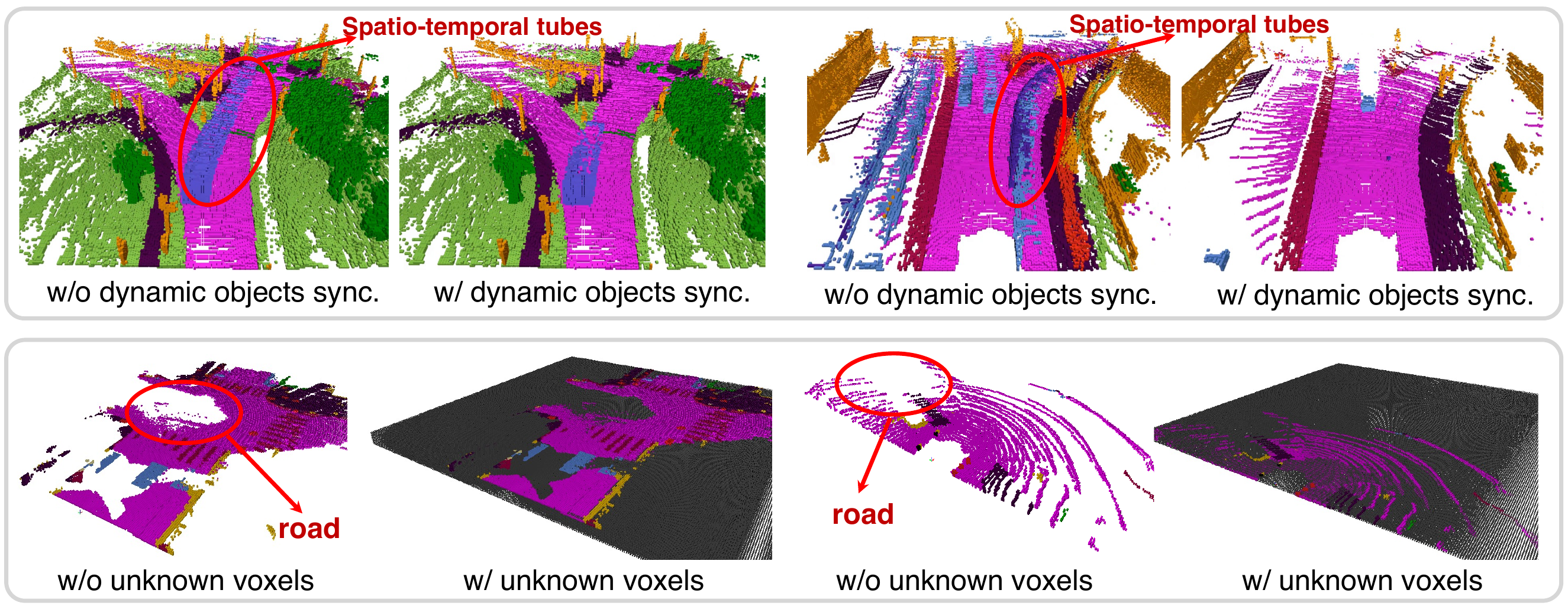}
   \vspace{-2mm}
   \caption{\textbf{Top row: dynamic objects synchronization.} Two examples on nuScenes~\citep{caesar2020nuscenes} are shown. Spatio-temporal tubes are introduced without handling dynamic objects, damaging the accuracy of labels. \textbf{Bottom row: unknown voxels exclusion.} Voxels are marked as unknown (denoted by grey color) when they are occluded or remain unprobed by the LiDAR. }
   \label{fig:dynamic-invalid}
   \vspace{-3.5mm}
\end{figure}

SemanticKITTI has two limitations. First, rigid registration with sensor poses can only handle measurements for static environments, resulting in traces produced by dynamic objects such as moving cars as shown in Fig.~\ref{fig:dynamic-invalid}, which can confuse the 3D representation learning~\citep{rist2021semantic}. Secondly, it is constrained by the limited scale and lack of diverse geographical coverage. The data collection is confined to a single city, resulting in training, validation, and test sets composed of 3,834, 815, and 3,992 frames, respectively, amounting to a total of 8,641 frames. However, this falls short of the large-scale benchmark necessary for comprehensive evaluation and generalization.

\subsection{SSCBench} We aim to establish a large-scale SSC benchmark in street views that facilitates the training of robust and generalizable SSC models. To achieve this, we harness well-established and widely-used datasets and integrate them into a unified setup and format. Overall, our SSCBench, consisting of three subsets, includes 38,562 frames for training, 15,798 frames for validation, and 12,553 frames for testing respectively, amounting totally to 66,913 frames ($\sim$67K), which greatly exceeds the scale of SemanticKITTI mentioned above by $\sim$7.7 times.
In the following, we introduce three carefully designed datasets, all based on existing data sources, that collectively contribute to our SSCBench.

\textbf{SSCBench-KITTI-360.}
KITTI-360~\citep{liao2022kitti} represents a significant advancement in autonomous driving research, building upon the renowned KITTI dataset~\citep{geiger2012we}. It introduces an enriched data collection framework with diverse sensor modalities as well as panoramic viewpoints (a perspective stereo camera plus a pair of fisheye cameras) and provides comprehensive annotations including consistent 2D and 3D semantic instance labels as well as 3D bounding primitives. The dense and coherent labels not only support established tasks such as segmentation and detection but also enable novel applications like semantic SLAM~\citep{bowman2017probabilistic} and novel view synthesis~\citep{zhang2023nerflets}. While KITTI-360 includes point cloud-based semantic scene completion, the prevalent methodology for SSC remains centered around voxelized representations~\citep{roldao20223d}, which exhibit broader applicability in robotics. 

\textit{\textbf{Remark.}} KITTI-360 covers a driving distance of 73.7km, comprising 300K images and 80K laser scans. While adhering to KITTI's forward-facing camera setup, it offers greater geographical diversity and demonstrates minimal trajectory overlap with KITTI. Leveraging the open-source training and validation set, we build SSCBench-KITTI-360 consisting of 9 long sequences. To reduce redundancy, we sample every 5 frames following the SemanticKITTI SSC benchmark. The training set includes 8,487 frames from scenes 00, 02-05, 07, and 10, while the validation set comprises 1,812 frames from scene 06. The testing set comprises 2,566 frames from scene 09. In total, the dataset contains 12,865 ($\sim$13K) frames, surpassing the scale of SemanticKITTI by $\sim$1.5 times. 

\textbf{SSCBench-nuScenes.}
Unlike KITTI's forward-facing camera setup, nuScenes~\citep{caesar2020nuscenes} captures a complete 360-degree view around the ego vehicle. It provides a diverse range of multimodal sensory data, including camera images, LiDAR point clouds, and radar data, gathered in Boston and Singapore. nuScenes offers meticulous annotations for complex urban driving scenarios, including diverse weather conditions, construction zones, and varying illumination. Later on, panoptic nuScenes~\citep{fong2022panoptic} extends the original nuScenes dataset with semantic and instance labels. With comprehensive metrics and evaluation protocols, nuScenes is widely employed in autonomous driving research~\citep{gu2023vip3d, hu2023planning,li2021fooling,huang2023tri}. 

\textit{\textbf{Remark.}} The nuScenes dataset consists of 1K 20-second scenes with labels provided only for the training and validation set, totaling 850 scenes. From the available 850 scenes, we allocate 500 scenes for training, 200 for validation, and 150 for testing. This distribution results in 20,064 frames for training, 8,050 frames for validation, and 5,949 for testing, totaling 34,078 frames ($\sim$34K). This scale is approximately four times that of SemanticKITTI. As nuScenes only provides annotations for keyframes at a frequency of 2Hz, there is no downsampling in SSCBench-nuScenes. 

\textbf{SSCBench-Waymo.}
The Waymo dataset~\citep{sun2020scalability}, collected from various locations in the US, offers a large-scale collection of multimodal sensor recordings. Waymo provides 5 cameras with a combined horizontal field of view of $\sim$230 degrees, slightly smaller than nuScenes. The data is captured in diverse conditions across multiple cities, including San Francisco, Phoenix, and Mountain View, ensuring broad geographical coverage within each city. It consists of 1000 scenes for training and validation, as well as 150 scenes for testing, with each scene spanning 20 seconds.

\textit{\textbf{Remark.}} To construct SSCBench-Waymo, we utilize the open-source training and validation scenes and redistribute them into sets of 500, 298, and 202 scenes for training, validation, and testing, respectively. We use only the annotated keyframes, which results in a training set of 14,943 frames, a validation set of 8,778 frames, and a test set of 5,946 frames, totaling 29,667 frames ($\sim$30K).

\subsection{Construction Pipeline}
\textbf{Prerequisites.} 
To establish SSCBench, a driving dataset with multimodal recordings is required for LiDAR-based or camera-based SSC. The dataset should include sequentially collected 3D LiDAR point clouds with accurate sensor poses for geometry completion, per-point semantic annotations for semantic scene understanding, and 3D bounding annotations to handle dynamic instances. 

\textbf{Aggregation of Point Clouds.} 
To generate a complete representation, our approach involves superimposing an extensive set of laser scans within a defined region in front of the vehicle. In short sequences like nuScenes and Waymo, we utilize future scans with measurements from the corresponding region to create a dense semantic point cloud. In long sequences like KITTI-360, which feature multiple loop closures, we incorporate all spatial neighboring point clouds in addition to the temporal neighborhood. Accurate sensor poses, provided by advanced SLAM systems~\citep{bailey2006simultaneous}, greatly facilitate the aggregation of point clouds for the static environment. As for dynamic objects, we avoid the spatial-temporal tubes by synchronization. We utilize the instance label to transform dynamic objects to their spatial alignment within the current frame. As shown in Fig.~\ref{fig:dynamic-invalid}, the spatial-temporal tubes are removed and objects have a denser shape.

\begin{figure}[t]
  \centering
   \includegraphics[width=0.95\linewidth]{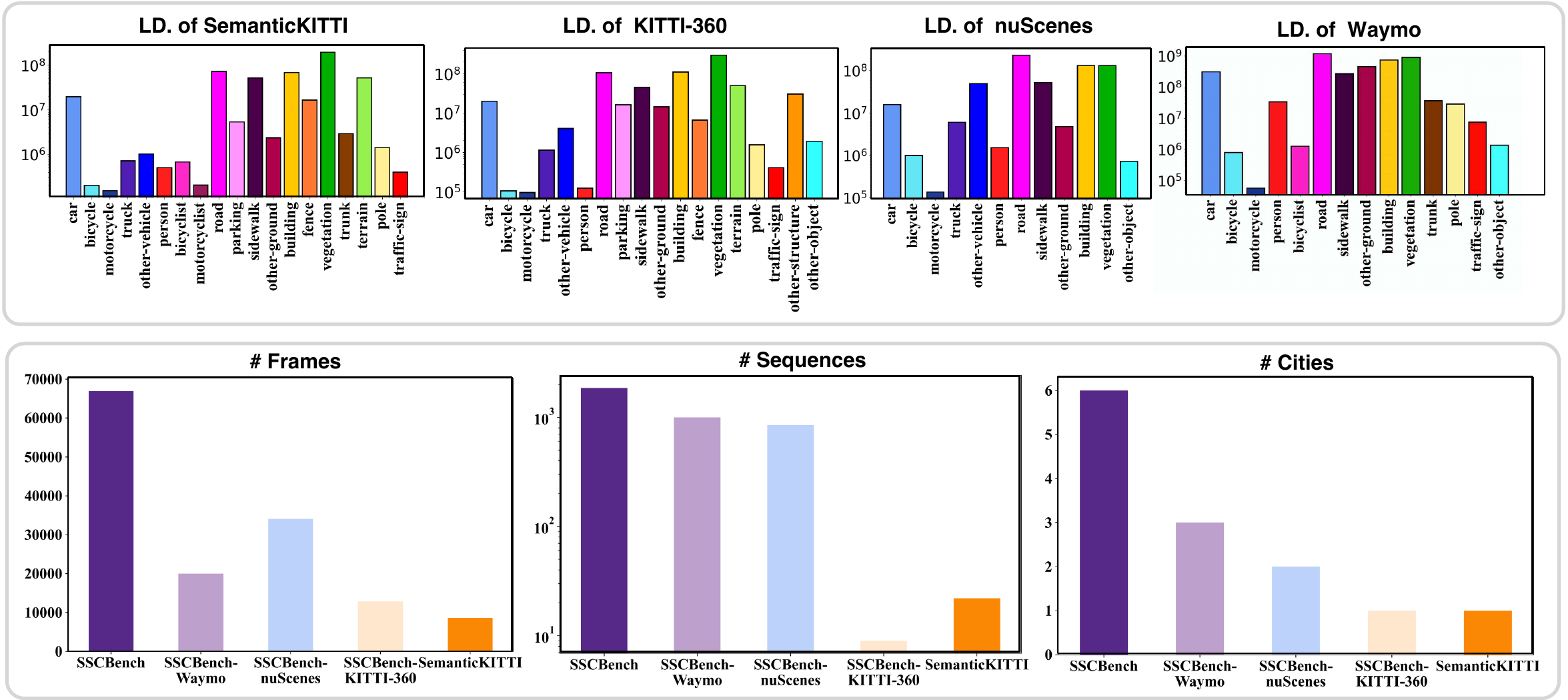}
   \vspace{-2mm}
   \caption{\textbf{Statistical analysis.} Top row: Label Distribution (LD.) of different datasets. Bottom row: scale comparisons between SSCBench and SemanticKITTI.}
   \label{fig:statis}
   \vspace{-4mm}
\end{figure}

\textbf{Voxelization of Aggregated Point Clouds.} 
Voxelization is to discretize a continuous 3D space into a regular grid structure composed of volumetric elements called voxels, enabling the conversion of unstructured data into a structured format that can be efficiently processed by convolutional neural networks (CNNs). Voxelization introduces a trade-off between spatial resolution and memory consumption and offers a flexible and scalable representation for 3D perception~\citep{maturana2015voxnet, zhou2018voxelnet,li2023voxformer}. For easy integration, SSCBench adheres to SemanticKITTI's setup, with a volume extending 51.2m ahead, 25.6m on each side, and 6.4m in height. Voxel resolution is 0.2m, resulting in a 256$\times$256$\times$32 voxel volume. Labels for each voxel are determined by majority voting among labeled points within it, while empty voxels are marked accordingly if no points are present. 

\textbf{Exclusion of Unknown Voxels.} 
Capturing complete 3D outdoor scenes is nearly impossible without ubiquitous scene sensing. While it is possible to utilize spatial or prior knowledge-based inference, we intend to ensure the fidelity of ground truth by minimizing errors originating from these steps. Hence, we only consider \textit{visible and probed} voxels from all viewpoints during training and evaluation. Specifically, we first employ ray tracing from different perspectives to identify and remove occluded voxels within objects or behind walls. Furthermore, in datasets with sparse sensing, where numerous voxels remain unprobed, we remove these unknown voxels during training and evaluation to enhance the reliability of the ground truth as shown in Fig.~\ref{fig:dynamic-invalid}.

\subsection{Dataset Statistics}
We show dataset statistics in Fig.~\ref{fig:statis}. The dataset label distribution reveals noticeable domain gaps among various cities. Specifically, KITTI-360 and SemanticKITTI exhibit similar label distributions due to being captured within the same city in Germany. However, nuScenes and Waymo collected in the US and Singapore demonstrate distinct label distributions. Furthermore, SSCBench stands out for its larger scale in comparison to SemanticKITTI. For instance, SSCBench comprises 7.7 times more frames than SemanticKITTI, and its collection of driving sequences is also more diverse.

\section{EXPERIMENTAL SETUP}

\label{sec:experimental setup}
\textbf{Benchmark Methods.} Our benchmark employs four camera-based methods, \textit{i.e.}, MonoScene~\citep{cao2022monoscene}, VoxFormer~\citep{li2023voxformer}, TPVFormer~\citep{huang2023tri}, and OccFormer~\citep{zhang2023occformer}, as well as two LiDAR-based methods, \textit{i.e.}, SSCNet~\citep{song2017semantic} and LMSCNet~\citep{roldao2020lmscnet}, due to their widespread adoption and good performance. Using their public codebases, we run them under default settings but appropriately adjust data loaders and batch sizes to align with our SSCBench. We train, validate, and test these methods separately on SSCBench subsets, as reported in Sec.~\ref{sec:individual_results}. Furthermore, we provide a unified benchmark for evaluating cross-domain generalizability in Sec.~\ref{sec:unfied_results}, where models are trained on one subset and tested on others, \textit{e.g.}, trained on SSCBench-KITTI-360 and tested on SSCBench-Waymo.


\textbf{Evaluation Metrics.} We adopt the intersection over union (IoU) for evaluating geometry completion and the mean IoU (mIoU) of each class for evaluating semantic segmentation, following SemanticKITTI. We also report the ratio of different classes in the dataset to better understand the relationship between IoU and mIoU. We report the performances within the volume extending 51.2m ahead, 25.6m on each side, and 6.4m in height, and the voxel resolution is 0.2m. The design of this front-view evaluation emphasizes the area directly in the vehicle’s anticipated forward trajectory.  


\section{Separate Benchmarking Results}\label{sec:individual_results}

\begin{table*}[t]\centering
\caption{\textbf{Separate benchmarking results on three SSCBench subsets.} We benchmark models with \textit{\textbf{monocular images}} and \textit{\textbf{point cloud}} inputs. The default evaluation range is 51.2$\times$51.2$\times$6.4m$^3$. Due to the label differences amongst the three subsets, missing labels are replaced with "-". The top three performances on each dataset are marked by \tbr{red}, \tbg{green}, and \tbb{blue} respectively.}
\label{tab:overall}
\renewcommand\tabcolsep{1.2pt}
\tiny
\resizebox{0.8\linewidth}{!}{
\begin{tabular}{c|c|c|c|c| c c c c c c c c c c c c c c c c c c c c}
\toprule


\textbf{\rotatebox{90}{Dataset}} &\textbf{\rotatebox{90}{Method}} &\textbf{\rotatebox{90}{Input}}  &\textbf{\rotatebox{90}{IoU}} &\textbf{\rotatebox{90}{mIoU}}
      &\rotatebox{90}{\crule[carcolor]{0.13cm}{0.13cm} \textbf{car}}
      &\rotatebox{90}{\crule[bicyclecolor]{0.13cm}{0.13cm} \textbf{bicycle}}
      &\rotatebox{90}{\crule[motorcyclecolor]{0.13cm}{0.13cm} \textbf{motorcycle}}
      &\rotatebox{90}{\crule[truckcolor]{0.13cm}{0.13cm} \textbf{truck}}
      &\rotatebox{90}{\crule[othervehiclecolor]{0.13cm}{0.13cm} \textbf{other-veh.}}
      &\rotatebox{90}{\crule[personcolor]{0.13cm}{0.13cm} \textbf{person}}
      &\rotatebox{90}{\crule[roadcolor]{0.13cm}{0.13cm} \textbf{road}}  
      &\rotatebox{90}{\crule[parkingcolor]{0.13cm}{0.13cm} \textbf{parking}}
      &\rotatebox{90}{\crule[sidewalkcolor]{0.13cm}{0.13cm} \textbf{sidewalk}}
      &\rotatebox{90}{\crule[othergroundcolor]{0.13cm}{0.13cm} \textbf{other-grnd}}
      &\rotatebox{90}{\crule[buildingcolor]{0.13cm}{0.13cm} \textbf{building}}
      &\rotatebox{90}{\crule[fencecolor]{0.13cm}{0.13cm} \textbf{fence}}
      &\rotatebox{90}{\crule[vegetationcolor]{0.13cm}{0.13cm} \textbf{vegetation}}
      &\rotatebox{90}{\crule[terraincolor]{0.13cm}{0.13cm} \textbf{terrain}}
      &\rotatebox{90}{\crule[polecolor]{0.13cm}{0.13cm} \textbf{pole}}
      &\rotatebox{90}{\crule[trafficsigncolor]{0.13cm}{0.13cm} \textbf{traf.-sign}}
      &\rotatebox{90}{\crule[other-struct.color]{0.13cm}{0.13cm} \textbf{other-struct.}}
      &\rotatebox{90}{\crule[other-objectcolor]{0.13cm}{0.13cm} \textbf{other-object}}
      &\rotatebox{90}{\crule[bicyclistcolor]{0.13cm}{0.13cm} \textbf{bicyclist}}
      &\rotatebox{90}{\crule[trunkcolor]{0.13cm}{0.13cm} \textbf{trunk}}
     
\\\midrule

\multirow{6}*{\rotatebox{90}{\parbox[t]{1cm}{SSCBench-\\KITTI-360}}} 
            & LMSCNet & L & \tbg{47.53} & \tbb{13.65} & {20.91} & {0} & {0} & {0.26} & {0} & {0} & \tbg{62.95} & \tbg{13.51} & \tbg{33.51} & {0.2} & \tbg{43.67} & {0.33} & \tbg{40.01} & \tbg{26.80} & {0} & {0} & {3.63} & {0.0003} & {-} & {-} 
        \\ & SSCNet & L & \tbr{53.58} & \tbr{16.95} & \tbr{31.95} & {0} & {0.17} & \tbr{10.29} & {0.58} & {0.07} & \tbr{65.7} & \tbr{17.33} & \tbr{41.24} & \tbb{3.22} & \tbr{44.41} & \tbr{6.77} & \tbr{43.72} & \tbr{28.87} & {0.78} & {0.75} & \tbr{8.60} & {0.67} & {-} & {-}
        \\ & MonoScene & C & {37.87} & {12.31} & {19.34} & {0.43} & \tbb{0.58} & {8.02} & {2.03} & {0.86} & {48.35} & {11.38} & {28.13} & \tbb{3.22} & {32.89} & {3.53} & {26.15} & {16.75} & \tbb{6.92} & {5.67} & {4.20} & \tbg{3.09} & {-} & {-}
        \\ & Voxformer & C & {38.76} & {11.91} & {17.84} & \tbr{1.16} & \tbg{0.89}& {4.56} & \tbb{2.06}  & \tbb{1.63} & {47.01} & {9.67} & {27.21} & {2.89} & {31.18} & \tbg{4.97} & {28.99} & {14.69} & {6.51} & \tbg{6.92} & {3.79} & {2.43} & {-} & {-}
        \\ & TPVFormer & C & {40.22} & {13.64} & \tbb{21.56} & \tbg{1.09} & \tbr{1.37} & \tbb{8.06} & \tbg{2.57} & \tbg{2.38} & {52.99} & {11.99} & {31.07} & \tbr{3.78} & {34.83} & \tbb{4.80} & {30.08} & {17.51} & \tbg{7.46} & \tbb{5.86} & \tbb{5.48} & \tbb{2.70} & {-} & {-} 
        \\ & OccFormer & C & \tbb{40.27} & \tbg{13.81} & \tbg{22.58} & \tbb{0.66} & {0.26} & \tbg{9.89} & \tbr{3.82} & \tbr{2.77} & \tbb{54.3} & \tbb{13.44} & \tbb{31.53} & \tbg{3.55} & \tbb{36.42} & \tbb{4.80} & \tbb{31.00} & \tbb{19.51} & \tbr{7.77} & \tbr{8.51} & \tbg{6.95} & \tbr{4.60} & {-} & {-} 
 
\\\midrule

\multirow{5}*{\rotatebox{90}{\parbox[t]{1cm}{SSCBench-\\nuScenes}}} 
            & LMSCNet & L & {21.09} & \tbb{8.36} & \tbg{14.74} & {0} & {0} & {5.93} & {8.52} & {3.41} & {24.14} & {-} & {7.55} & {8.40} & \tbg{18.56} & {-} & \tbg{9.02} & {-} & {-} & {-} & {-} & {0} & {-} & {-} 
        \\ & SSCNet & L & \tbb{27.64} & \tbr{11.84} & \tbr{18.06} & {0} & {0.36} & \tbr{13.42} & \tbr{10.35} & \tbr{7.59} & \tbb{28.74} & {-}& \tbb{12.65} & \tbg{12.65} & \tbr{21.05} & {-} & \tbr{16.33} & {-}  & {-} & {-} & {-} & \tbb{0.01} & {-} & {-}
        \\ & MonoScene & C & \tbr{29.63} & \tbg{9.60} & {10.17} & \tbg{1.7} & \tbg{3.80} & \tbb{8.35} & \tbb{8.74} & \tbb{3.72} & \tbr{38.77} & {-} & \tbg{14.74} & \tbb{12.58} & {7.23} & {-} & {5.50} & {-} & {-} & {-} & {-} & \tbg{0.03} & {-} & {-}
        \\ & Voxformer & C & {25.16} & {4.96} & {4.95} & \tbb{0.29} & \tbb{1.21} & {2.73} & {2.45} & {1.12} & {23.94} & {-} & {10.14} & {4.06} & {3.97} & {-} & {4.58} & {-} & {-} & {-} & {-} & \tbr{0.06} & {-} & {-}
        \\ & OccFormer & C & \tbg{28.23} & {7.55} & \tbb{14.61} & \tbr{2.25} & \tbr{7.97} & \tbg{11.88} & \tbg{9.80} & \tbg{5.87} & \tbg{37.62} & {-} & \tbr{18.63} & \tbr{19.76} & \tbb{9.05} & {-} & \tbb{5.92} & {-} & {-} & {-} & {-} & {0} & {-} & {-}
\\\midrule
\multirow{4}*{\rotatebox{90}{\parbox[c]{0.88cm}{SSCBench-\\Waymo}}}
            & LMSCNet & L & \tbr{71.62} & \tbg{36.29} & \tbr{75.63} & {0} & {0} & {-} & {-} & \tbr{67.08} & \tbg{71.79} & {-} & \tbg{40.67} & \tbg{41.57} & \tbr{51.40} & {-} & \tbg{49.40} & {-} & \tbg{33.82} & \tbg{32.64} & {-} & \tbg{2.36} & \tbg{16.28} & \tbg{25.46} 
        \\ & SSCNet & L & \tbg{70.44} & \tbr{39.10} & \tbg{71.34} & \tbg{1.78} & {0} & {-} & {-} & \tbg{64.72} & \tbr{72.44} & {-} & \tbr{46.47} & \tbr{46.11} & \tbg{51.15} & {-} & \tbr{49.47} & {-} & \tbr{34.72} & \tbr{39.18} & {-} & \tbr{19.72} & \tbr{19.56} & \tbr{30.68}
        \\ & MonoScene & C & \tbb{38.28} & \tbb{12.41} & \tbb{17.45} & \tbr{3.54} & \tbr{0.12} & {-} & {-} & \tbb{7.05} & {49.86} & {-} & {24.57} & {22.46} & {14.35} & {-} & {11.29} & {-} & \tbb{4.19} & \tbb{4.34} & {-} & \tbb{6.11} & \tbb{5.43} & \tbb{2.96}
        \\ & TPVFormer & C & {37.44} & {11.68} & {16.53} & \tbb{1.18} & \tbg{0.01} & {-} & {-} & {4.83} & \tbb{50.54} & {-} & \tbb{25.08} & \tbb{22.81} & \tbb{14.51} & {-} & \tbb{11.32} & {-} & {3.43} & {2.71} & {-} & {5.00} & {2.96} & {2.66}

\\\bottomrule
\end{tabular}
}
\vspace{-4mm}
\end{table*}

\subsection{Quantitative Comparisons}
\textbf{Camera-based Methods.} 
On SSCBench-KITTI-360, TPVFormer and OccFormer demonstrate superior geometry completion performance (IoU) compared to VoxFormer and MonoScene, as illustrated in Tab.~\ref{tab:overall}. This improved geometry completion also contributes to enhancing semantic segmentation (mIoU). 
However, on SSCBench-Waymo, MonoScene slightly outperforms TPVFormer in both geometry completion and semantic segmentation. On SSCBench-nuScenes, the IoU and mIoU metrics for MonoScene significantly surpass those of VoxFormer (IoU, $29.63 \rightarrow 25.16$ and mIoU, $9.60 \rightarrow 4.96$). Due to the absence of stereo data in SSCBench-nuScenes, the utilization of off-the-shelf self-supervised depth estimation modules~\citep{bhat2021adabins,shamsafar2022mobilestereonet}, primarily trained on KITTI, results in suboptimal depth knowledge, leading to less competitive performance by VoxFormer in SSC. It is evident that accurate depth estimation plays a crucial role in scene geometry estimation within camera-based methods.

\textbf{LiDAR-based Methods.} 
SSCNet consistently outperforms LMSCNet across all three subsets, mainly due to its larger number of parameters (1.03M compared to 0.35M). Additionally, it is worth noting that SSCNet exhibits superior recognition capabilities for small objects, such as motorcycles (\crule[motorcyclecolor]{0.20cm}{0.20cm},~$2.24 \leftrightarrow 0.00$ in SSCBench-nuScenes) and bicycles (\crule[bicyclecolor]{0.20cm}{0.20cm},~$1.78 \leftrightarrow 0.00$ in SSCBench-Waymo). This demonstrates SSCNet's advantage in handling sparse LiDAR data compared to LMSCNet. When comparing results between SSCBench-KITTI-360 and SSCBench-nuScenes to SSCBench-Waymo, SSCNet and LMSCNet consistently deliver significantly better performance on the SSCBench-Waymo dataset, with IoU values approaching 70\%. This improvement can be attributed to the denser LiDAR input available in Waymo data. However, it is crucial to emphasize that while dense LiDAR input leads to satisfactory performance, implementing this 5-LiDAR setup remains costly for common autonomous driving solutions.

\textbf{Camera \textit{vs.} LiDAR.} As demonstrated in Tab.~\ref{tab:overall}, on SSCBench-KITTI-360, LiDAR-based methods outperform camera-based approaches in terms of geometry metrics and most semantic metrics. This outcome is expected since camera-based methods must infer 3D scene geometry from 2D images, while LiDAR-based methods directly extract scene geometry from LiDAR input. However, the scenario changes in SSCBench-nuScenes, where camera-based methods generally surpass LiDAR-based methods in terms of IoU. This difference can be attributed to the use of a sparse LiDAR sensor (Velodyne HDL32E) in the nuScenes dataset. These results indicate that LiDAR-based methods are sensitive to the sparsity of input. Specifically, while dense input has the potential for significant performance improvement, sparse input can lead to significant degradation. This observation is further confirmed in SSCBench-Waymo, where the Waymo dataset contains point cloud data collected from one mid-range and four short-range LiDARs. As seen in Tab.~\ref{tab:overall}, the two LiDAR-based methods outperform camera-based methods by a significant margin on all metrics in SSCBench-Waymo.

However, camera-based methods outperform LiDAR-based ones for smaller objects that comprise a minuscule fraction of samples ($<0.5\%$). For classes such as bicycles (\crule[bicyclecolor]{0.2cm}{0.2cm}, $0.00 \rightarrow 1.16$), persons (\crule[personcolor]{0.2cm}{0.2cm}, $0.26 \rightarrow 4.54$), poles (\crule[polecolor]{0.2cm}{0.2cm}, $1.09 \rightarrow 12.93$), and traffic signs (\crule[trafficsigncolor]{0.2cm}{0.2cm}, $0.90 \rightarrow 14.25$) in SSCBench-KITTI-360, as well as motorcycles (\crule[motorcyclecolor]{0.2cm}{0.2cm}, $0.00 \rightarrow 3.80$) and bicycles (\crule[bicyclecolor]{0.2cm}{0.2cm}, $0.00 \rightarrow 1.70$) in SSCBench-nuScenes, camera-based methods significantly outperform LiDAR-based methods. Despite the low frequency of small objects, their identification is vitally important for collision avoidance and traffic understanding.

\begin{table}[t]\centering
\caption{\textbf{Comparison between monocular and trinocular setup on SSCBench-Waymo}. The monocular setup utilizes the front camera only. The trinocular setup utilizes images from all 3 front-facing cameras, which collectively provide a 180$^{\circ}$ view. }
\label{tab:tpvformer-mono-multi}
\renewcommand\tabcolsep{2pt}
\scriptsize
\resizebox{\columnwidth}{!}{%
\begin{tabular}{c|c|c|c| c c c c c c c c c c c c c c c}
\toprule


\textbf{\rotatebox{90}{Method}} &\textbf{\rotatebox{90}{Camera Setting} }&\textbf{\rotatebox{90}{IoU}} &\textbf{\rotatebox{90}{mIoU}}
      &\textbf{\rotatebox{90}{\crule[carcolor]{0.13cm}{0.13cm} car}}
      &\textbf{\rotatebox{90}{\crule[bicyclecolor]{0.13cm}{0.13cm} bicycle}}
      &\textbf{\rotatebox{90}{\crule[motorcyclecolor]{0.13cm}{0.13cm} motorcycle}}
      &\textbf{\rotatebox{90}{\crule[personcolor]{0.13cm}{0.13cm} person}}
      &\textbf{\rotatebox{90}{\crule[bicyclistcolor]{0.13cm}{0.13cm} bicyclist}}
      &\textbf{\rotatebox{90}{\crule[roadcolor]{0.13cm}{0.13cm} road}  }
      &\textbf{\rotatebox{90}{\crule[sidewalkcolor]{0.13cm}{0.13cm} sidewalk}}
      &\textbf{\rotatebox{90}{\crule[othergroundcolor]{0.13cm}{0.13cm} other-grnd}}
      &\textbf{\rotatebox{90}{\crule[buildingcolor]{0.13cm}{0.13cm} building}}
      &\textbf{\rotatebox{90}{\crule[vegetationcolor]{0.13cm}{0.13cm} vegetation}}
      &\textbf{\rotatebox{90}{\crule[trunkcolor]{0.13cm}{0.13cm} trunk}}
     &\textbf{\rotatebox{90}{\crule[polecolor]{0.13cm}{0.13cm} pole}}
      &\textbf{\rotatebox{90}{\crule[trafficsigncolor]{0.13cm}{0.13cm} traf.-sign}}
      &\textbf{\rotatebox{90}{\crule[other-objectcolor]{0.13cm}{0.13cm} other-object}}

\\\midrule

\multirow{2}*{TPVFormer}
    & Monocular
        & {37.44} & {11.68}& {16.53} & {1.18} & {0.01} & {4.83} & \textbf{2.95} & {50.54} & {25.08} & {22.80} & {14.51} & {11.32} & {2.66}& {3.43} &{2.71}& \textbf{5.00} \\
    & Trinocular
        & \textbf{39.22} & \textbf{13.78}& \textbf{20.59} & \textbf{1.50} & \textbf{0.04} & \textbf{9.11} & {2.29} & \textbf{53.37} & \textbf{28.26} &\textbf {25.43} & \textbf{16.60} & \textbf{16.05}& \textbf{5.17} &\textbf{5.92}&\textbf{4.23}&{4.40}
\\\bottomrule
\end{tabular}
}
\vspace{-4mm}
\end{table}

\subsection{Discussions and Analyses}
\textbf{Impact of Point Cloud Density.} Our experiments illuminate the impact of LiDAR input density on model performance. In the SSCBench-nuScenes dataset, which features relatively sparse LiDAR input (32 channels), camera-based methods outperform LiDAR-based methods on geometric metrics. However, in the SSCBench-Waymo dataset, which benefits from dense LiDAR input (64 channels, 5 LiDARs), LiDAR-based methods vastly outperform camera-based methods. The sensitivity of LiDAR-based methods to input becomes evident, with advantages observed in dense input and notable performance degradation in sparse input. This highlights the need for future research in developing robust LiDAR-based methods that can mitigate degradation while capitalizing on the benefits.


\textbf{Monocular \textit{vs.} Trinocular.}  Table~\ref{tab:tpvformer-mono-multi} displays the performance of TPVFormer with monocular and trinocular input. While a trinocular setup offers a broader field of view that can help enhance overall performance in terms of both IoU ($37.44 \rightarrow 39.22$) and mIoU ($11.68 \rightarrow 13.78$), achieving excellent results using only a single camera remains a compelling academic challenge. There is still significant research value in developing monocular methods that can match the performance of models with panoramic views, as they are memory-efficient, computationally efficient, and easy to deploy.


\textbf{Comparison with SemanticKITTI.} We observe significant discrepancies when comparing our experimental results on SSCBench to those from SemanticKITTI~\citep{behley2019semantickitti} (for more details, we refer readers to VoxFormer~\citep{li2023voxformer}). While VoxFormer performs admirably well on SemanticKITTI in metrics such as IoU and mIoU, it faces challenges with the diversity of our SSCBench dataset. This challenge primarily arises from its depth estimation module's inability to generalize beyond SemanticKITTI. Furthermore, LMSCNet, which typically exhibits superior geometric performance compared to SSCNet on SemanticKITTI, demonstrates the opposite trend on SSCBench. These discrepancies underscore two essential points. First, they highlight the significance of SSCBench, which provides diverse and demanding real-world scenarios for comprehensive evaluations. Second, they emphasize the necessity for robust methods capable of maintaining high performance across various environments.

\begin{table}[t]\centering
\caption{\textbf{Cross-domain evaluation}. We report the experiment results of training and testing on different datasets using unified labels.}
\renewcommand\tabcolsep{3.5pt}
\scriptsize
\resizebox{\columnwidth}{!}{%
\begin{tabular}{l|c|c|c|c| c c c c c c c c c c}
\toprule


\textbf{\rotatebox{90}{\parbox[t]{1cm}{Training-\\dataset}}} &\textbf{\rotatebox{90}{\parbox[t]{1cm}{Testing-\\dataset}}} &\textbf{\rotatebox{90}{Method}}   &\textbf{\rotatebox{90}{IoU}} &\textbf{\rotatebox{90}{mIoU}}
      &\textbf{\rotatebox{90}{\crule[carcolor]{0.13cm}{0.13cm} vehicle}}
      &\textbf{\rotatebox{90}{\crule[bicyclecolor]{0.13cm}{0.13cm} bicycle}}
      &\textbf{\rotatebox{90}{\crule[motorcyclecolor]{0.13cm}{0.13cm} motorcycle}}
      &\textbf{\rotatebox{90}{\crule[personcolor]{0.13cm}{0.13cm} person}}
      &\textbf{\rotatebox{90}{\crule[roadcolor]{0.13cm}{0.13cm} road}  }
      &\textbf{\rotatebox{90}{\crule[sidewalkcolor]{0.13cm}{0.13cm} sidewalk}}
      &\textbf{\rotatebox{90}{\crule[othergroundcolor]{0.13cm}{0.13cm} other-grnd}}
      &\textbf{\rotatebox{90}{\crule[buildingcolor]{0.13cm}{0.13cm} building}}
      &\textbf{\rotatebox{90}{\crule[vegetationcolor]{0.13cm}{0.13cm} vegetation}}
      &\textbf{\rotatebox{90}{\crule[other-objectcolor]{0.13cm}{0.13cm} other-object}}
      
\\\midrule

\multirow{6}*{\rotatebox{90}{\parbox[t]{1.2cm}{SSCBench-\\KITTI-360}}}
    & \multirow{3}*{\parbox[t]{1.2cm}{SSCBench-\\KITTI-360}}
        &   LMSCNet
            & {48.49} &{22.48}& {23.87} & {0.00} & {0.00} & {0.00} & {64.04} & {35.13} & {17.46} & {44.47} & {39.755} & {0.04}  \\
        &&   SSCNet
            & \textbf{54.50}& \textbf{25.63} & \textbf{34.14} & {0.00} & {0.00} & {0.11} & \textbf{67.54} & \textbf{41.84} & \textbf{19.86} & \textbf{46.89} & \textbf{44.92} & {1.01} \\
        &&   MonoScene
            & {40.56} & {20.78}& {23.00} & \textbf{3.60} & \textbf{3.00} & \textbf{4.58} & {54.00} & {31.96} & {14.39} & {36.29} & {30.45} & \textbf{6.51}
    \\\cmidrule{2-15}
    & \multirow{3}*{\parbox[t]{1.2cm}{SSCBench-\\Waymo}} 
        &   LMSCNet
            & {0.11}& {0.02} & {0.38} & {0.00} & {0.00} & {0.00} & {0.31} & {0.07} & {0.06} & {10.69} & {10.27} & {0.00} \\

        &&   SSCNet
            & {0.13}& {0.04}& \textbf{6.35} & {0.00} & {0.00} & {0.00} & \textbf{4.02} & \textbf{2.31} & \textbf{1.49} & \textbf{15.56} & \textbf{11.44} & \textbf{0.62} \\
        &&   MonoScene
            & \textbf{7.16} & \textbf{2.14}& {4.89} & \textbf{1.05} & {0.00} & \textbf{0.67} & {1.15} & {0.81} & {0.58} & {4.85} & {6.84} & {0.55}
\\\cmidrule{1-15}

\multirow{6}*{\rotatebox{90}{\parbox[t]{1.2cm}{SSCBench-\\Waymo}}}
    & \multirow{3}*{\parbox[t]{1.2cm}{SSCBench-\\KITTI-360}}
        &   LMSCNet
            & {0.10}& {0.02} & {0.58} & {0.00} & {0.00} & {0.01} & {1.45} & {0.10} & {0.91}& {9.95} & {9.93} & {0.09} \\
        &&   SSCNet
            & {0.17}  & {0.05}& {2.73} & {0.00} & {0.00} & {0.04} & {7.19} & {1.39} & {1.29} & \textbf{14.67} & \textbf{17.86} & {0.07} \\
        &&   MonoScene
            & \textbf{16.29}& \textbf{3.99} & \textbf{6.55} & {0.15} & {0.01} & \textbf{0.80} & \textbf{7.76} & \textbf{2.93} & \textbf{3.14} & {6.95} & {11.30} & \textbf{0.34} 
    \\\cmidrule{2-15}
    & \multirow{3}*{\parbox[t]{1.2cm}{SSCBench-\\Waymo}} 
        &   LMSCNet
            & \textbf{71.77} & \textbf{43.57}& \textbf{77.17} & {0.00} & {0.00} & \textbf{68.78} & {70.28} & {37.65} & {42.79} & {51.63} & {49.52} & {37.89} \\

        &&   SSCNet
            & {71.00}& {43.48} & {69.68} & {0.28} & {0.00} & {63.30} & \textbf{70.92} & \textbf{42.73} & \textbf{46.27} & \textbf{51.49} & \textbf{50.38} & \textbf{39.69} \\
            
        &&   MonoScene
            & {37.79}& {15.57} & {17.84} & \textbf{3.32} & \textbf{0.01} & {7.64} & {49.16} & {23.09} & {23.33} & {13.40} & {13.44} & {4.45} 

\\\bottomrule
\end{tabular}
}
\label{tab:uni}
\vspace{-4mm}
\end{table}

\section{Unified Benchmarking Results}\label{sec:unfied_results}

To assess the domain gap and compare the cross-domain generalizability of state-of-the-art algorithms, we established a unified benchmark for cross-validation on SSCBench. Specifically, we employed two LiDAR-based methods, LMSCNet~\citep{roldao2020lmscnet} and SSCNet~\citep{song2017semantic}, and one camera-based method, MonoScene~\citep{cao2022monoscene}, for experiments on SSCBench-KITTI-360 and SSCBench-Waymo. To ensure consistent evaluation metrics, we standardized the labels of SSCBench-KITTI-360 and SSCBench-Waymo to a unified set comprising 10 common objects. All other experimental settings and evaluation metrics adhere to the guidelines outlined in Sec.~\ref{sec:experimental setup}.

\textbf{Overall Performance.} 
As shown in Tab.~\ref{tab:uni}, all three methods exhibit a notable decline in performance when cross-validated on another dataset across both the geometric metric (IoU) as well as the semantic metric (mIoU), regardless of the training dataset. Specifically, the model trained on SSCBench-Waymo and tested on SSCBench-KITTI-360 suffers a more severe decline for LiDAR-based methods than the other way around. This is because SSCBench-Waymo has a very dense point cloud input from five LiDARs, which effectively reduces the performance degradation caused by domain differences. Interestingly, the deterioration trend in terms of mIoU for MonoScene is more severe when transferring from SSCBench-KITTI-360 to SSCBench-Waymo than the other way around. This can be partially explained by the higher in-domain mIoU on SSCBench-KITTI-360 than that on SSCBench-Waymo and the difference in input resolutions (1408$\times$376 $\leftrightarrow$ 960$\times$640), which is magnified by the fixed model parameters, and thereby affects feature representation.

\textbf{Class-Specific Performance.} 
We observe the most significant performance drop in the "road" class (\crule[roadcolor]{0.2cm}{0.2cm}) for both transfer directions in all three methods. This suggests that ground types may be represented differently across datasets, causing challenges for both camera-based and LiDAR-based methods to adapt to these variations. 
Cross-domain evaluations consistently show a decline in performance, highlighting the importance of our proposed SSCBench dataset. Training models on this dataset should better prepare them for the variations and complexities of cross-domain scenarios. Moreover, it also serves as motivation for the development of models that are more robust and capable of generalizing across different domains.

\section{CONCLUSIONS}

\textbf{Limitations and Future Work.} SSCBench only encompasses 3D data following the convention of the SSC problem. This limits evaluations of 4D methods with temporal dimensions. Future work will aim to expand SSCBench to include temporal information.

\textbf{Summary.} In this paper, we introduce SSCBench, a large-scale benchmark composed of diverse street views, aimed at facilitating the development of robust and generalizable semantic scene completion models. Through meticulous curation and comprehensive benchmarking, we identify the bottlenecks of existing methods and offer valuable insights into future research directions. Our ambition is for SSCBench to stimulate advancements in 3D semantic scene completion, ultimately enhancing perception capabilities for the next-generation autonomous systems.

{\small
\bibliographystyle{IEEEtranN}
\bibliography{IEEEabrv,IEEEexample}}

\end{document}